# Accurate Autism Spectrum Disorder prediction using Support Vector Classifier based on Federated Learning (SVCFL)

Ali Mohammadifar[1], Hasan Samadbin[1], Arman Daliri[1]

[1] Department of Computer Engineering, Karaj Branch, Islamic Azad University,
Karaj, Iran

Ali.Mohammadifar@kiau.ac.ir
Hasan.Samadbin@Kiau.ac.ir
Arman.Daliri@kiau.ac.ir

*Abstract*

**The path to an autism diagnosis can be long and difficult, and delays can have serious consequences. Artificial intelligence can completely change the way autism is diagnosed, especially when it comes to situations where it is difficult to see the first signs of the disease. AI-based diagnostic tools may help confirm a diagnosis or highlight the need for further testing by analyzing large volumes of data and uncovering patterns that may not be immediately apparent to human evaluators. After a successful and timely diagnosis, autism can be treated through artificial intelligence using various methods. In this article, by using four datasets and gathering them with the federated learning method and diagnosing them with the support vector classifier method, the early diagnosis of this disorder has been discussed. In this method, we have achieved 99% accuracy for predicting autism spectrum disorder and we have achieved 13% improvement in the results.**

## 1. Introduction

Autism spectrum disorder (ASD) is a type of developmental disorder [1]. This disorder includes difficulties interacting and developing social relationships, difficulties with sensory processing, and repetitive behaviors [1]. The term "spectrum" reflects the fact that symptoms vary from person to person and vary in type and intensity [1]. There are many signs and symptoms that indicate that a person has autism, and of course, some adults and non-autistic children may also show a number of behaviors and symptoms of autism [2]. For example, some children who are not diagnosed with autism later may also walk on tiptoes for some reason, but this symptom decreases in them by the age of 18 months [2]. It is one of these symptoms that can be shared by non-autistic and autistic people [2].

The World Health Organization estimates that one in every 100 children worldwide has autism spectrum disorders [3]. According to the US Centers for Disease Control and Prevention's Autism and Developmental Disabilities Surveillance Network, approximately one in 44 American eight-year-olds was diagnosed with autism in 2018 [4]. Artificial intelligence has shown promising progress in the diagnosis and treatment of behavioral disorders such as autism in the past few years [5]. Gone are the days when diagnosing a child's autism required months or even years of observation, testing and evaluation by medical professionals [5].

In most cases of autism, delayed diagnosis limits the child's chances of success [6]. Imagine a society where people on the autism spectrum are quickly identified and effectively treated using artificial intelligence (AI) [6]. In this way, delays can be avoided by introducing diagnostic tools based on artificial intelligence [6]. Deep learning AI systems can identify patterns and behaviors that may be signs of autism, which can lead to faster and more accurate diagnosis [7]. It may mean that treatments and therapies are available sooner, enabling people to realize their full potential [7]. In many ways, the field of autism has revolutionized artificial intelligence rather than simple diagnosis.

In this article, an attempt has been made to improve the early prediction of autism spectrum disorder by using machine learning methods. Among the defects that exist in this type of prediction, issues such as: defects of databases, problems of lack of data, problems of autism questionnaires have been discussed. In this research, support vector classifier and federated learning methods have been used to solve these problems. In this research, four datasets have been integrated using federated learning method. Then, using the support vector classifier algorithm, the disease has been diagnosed. Among the main contributions of this article, the following can be mentioned:

1. Providing a framework for early prediction of autism spectrum disorder.
2. Combination of federated learning method and support vector classifier for accurate prediction.
3. Validated data analysis to combine with the federated learning method.

The structure of the article is as follows, firstly, the literature review is discussed in section 2. Then, in section 3, the proposed method is presented. In the fourth section, the results are reviewed and compared with other algorithms. Finally, section 5 presents the conclusion and future works are mentioned in it.

## 2. RELATED WORKS

In this section, three issues have been discussed. First, in sub-section A, the introduction of autism spectrum disorder has been discussed and the methods of understanding it by treatment clinics have been given. Then, in sub-section B, explanations about federated learning methods and its classification are discussed.

### A. About autism spectrum disorder and its identification

methods Autism spectrum disorder in DSM-5 is a general diagnosis that has replaced the four pervasive developmental disorders in the previous version: autism disorder, Asperger's disorder, childhood dissociative disorder, pervasive developmental disorder unspecified [8]. The chain or spectrum of ASD symptoms in individuals varies from mild to severe [9]. In the term, those who experience mild to moderate symptoms are said to have Asperger's symptoms. But there is no official category for diagnosis [9]. Autistic people seem apathetic and have difficulty connecting emotionally with others [10]. Their response to sensory experience is also unusual. For example, they will be distracted by the sound of a dripping faucet [10].

Autism spectrum disorder is seen in many countries and in racial, ethnic, religious and economic backgrounds [11]. It is estimated that almost one percent of the world's population has autism [11]. Research by the Centers for Disease Control and Prevention shows that one out of every 59 children in the United States have an autism spectrum disorder [11]. The sooner the disorder is diagnosed, the sooner the child can be helped through therapeutic measures [11]. Usually, the symptoms of autism spectrum disorder or ASD appear from the age of two, and doctors can diagnose the disorder at this age. In severe cases, symptoms may appear earlier [12]. While those whose symptoms of autism spectrum disorder are relatively mild until adolescence or adulthood may not notice it [12].

### B. About Federated Learning

Federated learning (also known as cooperative learning) is a machine learning method that trains an algorithm on multiple decentralized edge devices or servers that hold local data samples, without exchanging them [13]. Federated learning enables multiple actors to build a robust joint machine learning model without sharing data, thus allowing to address critical issues such as data privacy, data security, data access rights, and access to heterogeneous data [14]. The purpose of federated learning is to train a machine learning algorithm, i.e., deep neural networks, on multiple local datasets that exist at local nodes without explicitly exchanging data [15].

Generally, federated learning is divided into three categories: horizontal, vertical, and transfer federated learning [16]. One type of federated learning is horizontally federated, where the data distribution is similar across participants, but providers do not overlap to the same extent. In this type of federated learning, each machine learning model is identical and complete and can be predicted independently. As a result, this process can be placed as a distributed training on the scenario of vertical federated learning versus horizontal federated learning. In this type of learning, the user sets are all the same, but there are different data types from these users. For example, airlines and hotels have different data from the same user. Consequently, vertical federated learning requires sample alignment and model encoding. The strict transfer federated learning application scenario has a limited number of identical users and a very small dataset with identical characteristics [16]. Figure 1 presents the types of federated learning.

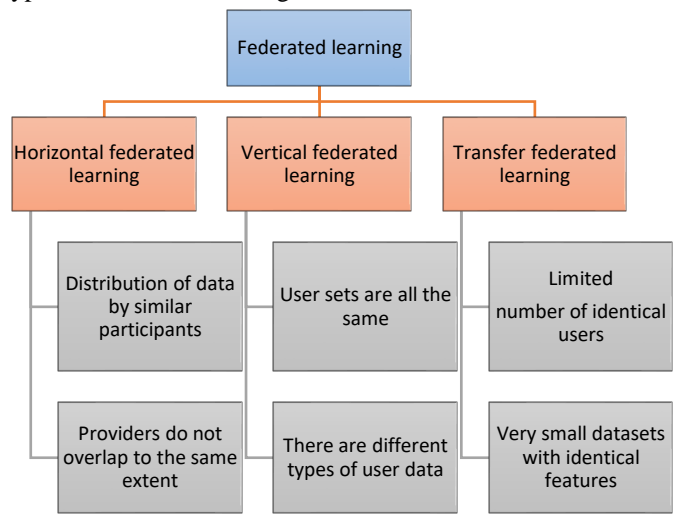

*Figure 1.Types of federated learning*

## 3. SUPPORT VECTOR CLASSIFIER BASED ON FEDERATED LEARNING (SVCAL)

In this section, the method proposed in this article is explained. First, the database used has been introduced. Then the main parts of the proposed framework have been explained. In general, this framework consists of two parts. First, in section A, the first part of the framework, which is the data preparation part, has been explained. Then in part B, the explanation of data collection with the help of federated learning has been discussed.

The advantage of our work is the application of FL technique for detection of ASD in both children and adults. Support Vector classifier model have been trained locally using four different ASD datasets of features containing records about children and adults obtained from free sources and data providing agencies listed in Table 1 for autism detection.

*Table 1. Dataset description of children and adults.*

| Categories | Source | instances | attributes |
|---|---|---|---|
| Children | https://archive.ics.uci.edu/ml/datasets/Autistic+Spectrum+Disorder+Screening+Data+for+Children++ https://www.kaggle.com/datasets/fabdelja/autism-screening-for-toddlers | 1,346 | 40 |
| Adults | https://archive.ics.uci.edu/ml/datasets/Autism+Screening+Adult https://www.kaggle.com/datasets/andrewmvd/autism-screening-on-adults | 1404 | 40 |

### A. Data preparation

The data used in this framework have several flaws that are not mentioned in the reference articles. Among the flaws of the data, it can be mentioned that two reference datasets are copies of four datasets. Also, many features are empty and cannot be filled with statistical methods. For this reason, many changes have been made in the data and we have arranged them in an integrated manner. Consistent with Q-chart 10, treating the adult and child datasets on the same scale, the 10 different highlights for separating extreme introverted patients from normal patients are consistent, as shown in Table 2, was identified.

### B. Federated learning

In this article, we proposed a special federated learning-based show that investigates four different datasets of adults and children in the field using SVC and prepares a demonstration of local information. These local models were transferred to a central server to create a worldwide meta-classifier for preventing extreme introversion in children and adults. The basis for carrying out this study was the screening approach "Quantitative Checklist for Extreme Introversion in Children" (Q-CHART-10), endorsed by the Changing Extreme Introverted Ness Venture in the United Kingdom.

A score above three means that the ASD highlighting is stopped, its weight is increased by 1, and "Yes" is moved to the response set. "No" is stored in the reaction sentence. Each price variable is compared to multiple questions to ensure that it is close to the values included in the Q-CHART-10 checklist. The data stored in the class answer set is in a parallel array and indicates "yes" (stored as 1) and "no" (stored as 0). Neighborhood ML models were built for these responses shown in Table 2.

The response data set contained an impulsive missing data set, which required data changes to be made earlier to prepare the ML classifier for demonstration preparation and research. Category elements are processed using name encoding. Name encoding converts names into numeric format to make them machine-readable. A warmed-up name is evaluated in the same way as one that is already named. Parallel name coding of classes with 10 emphases was chosen. The path of the proposed method is shown in Figure 2.

*Table 2. Feature description.*

| Feature | Title | Description | Response |
|---|---|---|---|
| 1 | Patient's age | Children (0–15 years), adults (16 years or above) | R1 |
| | Patient's gender | Male/female | |
| | Ethnicity of patient | Common ethnicities | |
| | Residence country | Countries list | |
| 2 | Verbal communication | Response on calling Name, saying papa, mama | R2 |
| 3 | Non-verbal communication | Eye contact, facial expressions, gestures, posture, use of objects and body language | R3 |
| 4 | Sensory processing | See, hear, smell, taste, touch | R4 |
| 5 | Repetitive behavior | Do action again and again | R5 |
| 6 | Motor skills | Walking, running, riding a bike | R6 |
| 7 | Preservative thinking | Rumination, repetitive thinking, worry | R7 |
| 8 | Having jaundice | Child/adult born with jaundice | R8 |
| 9 | Person taking ASD test | Parent staff, caregiver etc. | R9 |
| 10 | Used the screening app before | Did the user use a screening app for ASD | R10 |
| | Screening method type | Type of methods of screening chosen based on age category | |

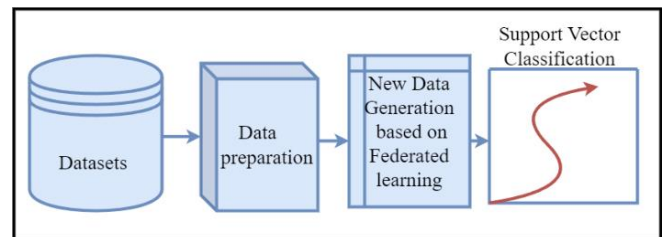

*Figure 2. SVCFL Graphical Abstract*

## 4. EXPERIMENTAL DESIGN

In this section, we report experiments to evaluate the proposed framework. However, the purpose of this task is to accurately classify her three answers to the class. In particular, we use decision tree (DT) [17], random forest (RF) [18], and support vector machine (SVM) [19] classifiers. All initial parameters for each classifier are set to default values.

Table 3 shows the evaluation metrics, F1 score, precision and recall of FL and the three classification methods. The rest of the table shows the classification performance when the FL method fits the data. As the results show, the balanced FL data classification provides better performance than the raw data. SVC has the highest F1 score compared to other classification methods. For other evaluation measures, this classifier's results are the closest fit. Note that some classifiers have higher evaluation metrics than SVC. However, the overall performance of this method for all questions is better than the others.

Note that the results are in two parts, FL which is after the implementation of the federated learning system and the other part is the implementation of the models using Raw Data. The effect of data collection method with FL is very effective. The accuracy results of DT, RF and SVC algorithms when the data were raw were 62%, 53% and 63%, respectively. After implementing the algorithms with the FL method, the algorithms had an accuracy of 94%, 94% and 99%. The reason for this amount of improvement in this method is the increase of examples and the optimization of features that Federated Learning has provided.

*Table 3. Evaluation metrics, F1 score, precision and recall of FL and the three classification methods*

| Method | Classifier | PR | R | Fs | Acc |
|---|---|---|---|---|---|
| Raw Data (Adults) | DT | 85% | 15% | 75% | 62% |
|  | RF | 83% | 49% | 81% | 53% |
|  | SVC | 86% | 52% | 82% | 63% |
| **Federated learning** | DT | 95% | 94% | 95% | 94% |
|  | RF | 95% | 94% | 95% | 94% |
|  | **SVC** | **99%** | **99%** | **99%** | **99%** |

## 5. CONCLUSION

In this article, a new method for predicting autism spectrum disorder is presented. This work has been implemented by analyzing several datasets and examining two different areas of machine learning. First, the analysis and preparation of the data was done, then the implementation of the federated learning method was done, and the support vector machine method was used for diagnosis. In this method, the prediction accuracy has reached 99% from 86% accuracy. Considering that a significant effort has been made in this method, future works can be complementary. For future works, it is suggested to try to implement transfer learning methods and use large datasets with higher accuracy. It is also possible to work with autism treatment centers for future work from data collection with the help of application software.